\documentclass[runningheads]{llncs}

\usepackage[T1]{fontenc}

\usepackage{graphicx}
\usepackage{subfig}

\begin{document}
\title{A Partially Reversible U-Net for Memory-Efficient Volumetric Image Segmentation}
\titlerunning{Partially Reversible, Memory-Efficient U-Net}
%
\author{Robin Br\"ugger \and
Christian F.\ Baumgartner \and
Ender Konukoglu}
\authorrunning{R. Br\"ugger et al.}
%
\institute{Computer Vision Lab, ETH Z\"urich, Switzerland
\email{\{baumgartner, ender.konukoglu\}@vision.ee.ethz.ch}}
\maketitle              
\begin{abstract}
One of the key drawbacks of 3D convolutional neural networks for segmentation is their memory footprint, which necessitates compromises in the network architecture in order to fit into a given memory budget. Motivated by the RevNet for image classification, we propose a partially reversible U-Net architecture that reduces memory consumption substantially. The reversible architecture allows us to exactly recover each layer's outputs from the subsequent layer's ones, eliminating the need to store activations for backpropagation. This alleviates the biggest memory bottleneck and enables very deep (theoretically infinitely deep) 3D architectures. On the BraTS challenge dataset, we demonstrate substantial memory savings. We further show that the freed memory can be used for processing the whole field-of-view (FOV) instead of patches. Increasing network depth led to higher segmentation accuracy while growing the memory footprint only by a very small fraction, thanks to the partially reversible architecture.

\end{abstract}
\section{Introduction}
The design of 3D segmentation networks is often severely limited by GPU memory consumption. The main issue is the training. The output of each layer (referred to as activations for the remainder of this paper) needs to be stored for the backward pass. This problem is exacerbated for deeper networks and when feeding the network with larger field-of-views. Alleviating this issue has the potential to yield accuracy gains, especially considering recent success of 3D architectures \cite{DBLP:journals/corr/abs-1809-10483,DBLP:journals/corr/abs-1810-11654}. While the memory problem can always be mitigated by better hardware, the proposed method is a more cost effective and sustainable solution. Furthermore, our method allows for the design of 3D segmentation networks with depths that might not have been possible before on any hardware.

Several strategies for reducing memory consumption of segmentation CNNs have been proposed in the literature. One possible strategy is processing the volume in patches. Kamnitsas et al.\ (DeepMedic) \cite{KAMNITSAS201761} used patches sampled at two different scales of size 25 x 25 x 25 and 19 x 19 x 19 voxels. They noted that contextual information from a large area is beneficial for accuracy. This is also evidenced by a trend towards bigger patch sizes. For instance, the winner of the BraTS 2018 challenge \cite{DBLP:journals/corr/abs-1810-11654} used a large  patch size of 160x192x128, which covers almost the entire FOV, but required a GPU with 32 GB of memory for training. 

A second method to reduce the memory consumption is to train with small batch sizes. For example, both the V-Net \cite{7785132} and the No-New-Net  \cite{DBLP:journals/corr/abs-1809-10483} used a batch size of two. The winner of the BraTS 2018 challenge even used a batch size of one \cite{DBLP:journals/corr/abs-1810-11654}. Small batch sizes do not work well with the commonly used Batch Normalization but Group Normalization (GN) can be used instead \cite{DBLP:journals/corr/abs-1803-08494}.

The potential memory savings of patch-based approaches and small batch sizes is limited. Small patch sizes lack global context, and the smallest possible batch size is 1. On the other hand, most of the memory during training of a 3D neural network is taken up by the need to store activations. Reducing the need to store activations can lead to substantial memory gains well beyond the other strategies. For a classification network as well as an RNN, Chen et al.\ \cite{DBLP:journals/corr/ChenXZG16} proposed to only store these values at certain intervals. When they are needed during back-propagation, the missing activations can be recomputed by an additional forward pass. Assuming a simple, nonbranching feed-forward network, storing the activations at every $\sqrt{n}$'th layer of a $n$-layer network reduces the memory requirement from $O(n)$ to $O(\sqrt{n})$ while doubling the computational cost of the forward passes. Recursively applying this technique allows for as little as $O(\log{n})$ memory for the activations.
\begin{figure}
	\centering
	\subfloat[Forward computation]{\includegraphics[width=5.5cm]{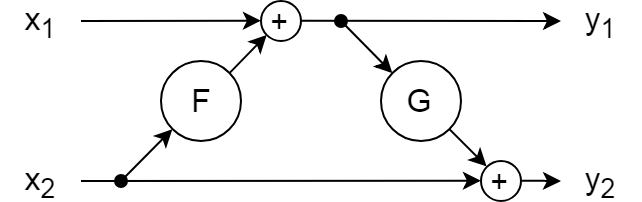}}
	\qquad
	\subfloat[Backward computation]{\includegraphics[width=5.5cm]{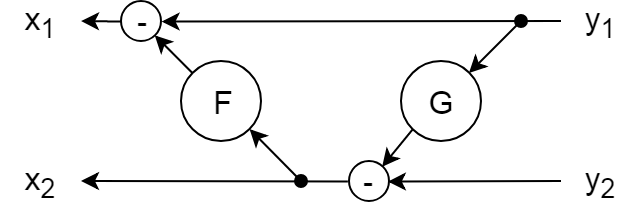}}
	\caption{Reversible block}
	\label{fig:reversible_block}
\end{figure}

Reversible layers proposed by Gomez et al.\ \cite{NIPS2017_6816} take the idea of not storing activations even further. During the backward pass, each layer's activations can be calculated from the following layer's activations. Therefore, no intermediate activations need to saved, allowing for $O(1)$ memory cost for the activations. However, for this to work the number of inputs of a layer needs to match the number of outputs.

The basic building element of a reversible neural network is the reversible block. It is illustrated in Fig.~\ref{fig:reversible_block}. In the forward computation, it takes inputs $x_1$ and $x_2$ of equal shape and outputs $y_1$ and $y_2$. The forward computation can be expressed with the following equations:
\begin{equation}
y_1 = x_1 + F(x_2) \qquad y_2 = x_2 + G(y_1) \label{eq:reversible_forward1}
\end{equation}
$F$ and $G$ can be any arbitrary function, as long as the output shape is the same as the input shape. An example for $F$ and $G$ is a convolutional layer that maintains channel size and number followed by an activation function. The reversible block partitions the layer's input into two groups, one flowing from the top path in Fig.~\ref{fig:reversible_block}(a) and one from the bottom. Among different partitioning strategies, Gomez et al.\ \cite{NIPS2017_6816} found that partitioning the channels works best. Following \cite{NIPS2017_6816}, we partition the channels of the activation maps into two groups for all reversible blocks in this work.

\begin{figure}
	\includegraphics[width=12cm]{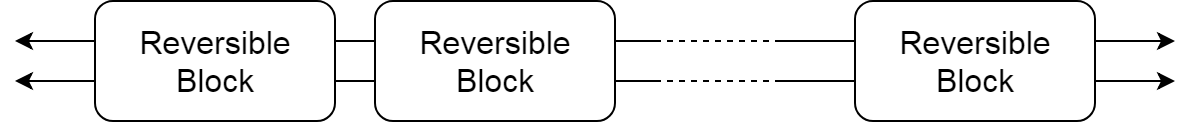}
	\caption{Reversible sequence}
	\label{fig:reversible_sequence}
\end{figure}

In the backward pass, the derivatives of the parameters of $F$ and $G$, as well as the reversible block's original inputs are calculated. The design of the reversible block allows to recover $x_1$ and $x_2$ given only $y_1$ and $y_2$ using Equation \ref{eq:reversible_backward1}, thus making it reversible.
\begin{equation}
x_2 = y_2 - G(y_1) \qquad x_1 = y_1 - F(x_2) \label{eq:reversible_backward1}
\end{equation}
Multiple reversible blocks can be chained together to form a sequence of arbitrary length. Because every block is reversible, the entire sequence is reversible as well. We refer to this as a \textit{reversible sequence} (RSeq) and illustrate it in Fig.~\ref{fig:reversible_sequence}. An RSeq is an important building block for our partially reversible architecture because the activations only have to be saved at the end of a RSeq, hence an entire RSeq has almost the same memory footprint as a single block with the difference being memory required to store extra weights, which is a very small fraction of the activations for 3D convolutional layers. The length of the sequence can therefore be varied (almost) free of memory constraints. That allows us to create a family of architectures which vary in the number of blocks per RSeq.

Blumberg et al.\ \cite{DIQT} used reversible blocks to create a very deep network for the task of super-resolution. They employed an architecture without skip connections and mainly focused on accuracy gains. We on the other hand propose a partially reversible U-Net with the main aim of enabling deep and powerful 3D segmentation networks on commodity hardware.

\section{Methods}
\subsection{Partially reversible architecture}
\label{section:architectures}
In this section, we describe our novel partially reversible U-Net architecture. We chose the No-New-Net by Isensee et al.\ \cite{DBLP:journals/corr/abs-1809-10483} as a non-reversible starting point and extend it. The No-New-Net won the second place in the BraTS 2018 challenge and was trained on a graphics card with 12GB of memory, a commonly used and widely available type of hardware. 

\begin{figure}
	\includegraphics[width=12cm]{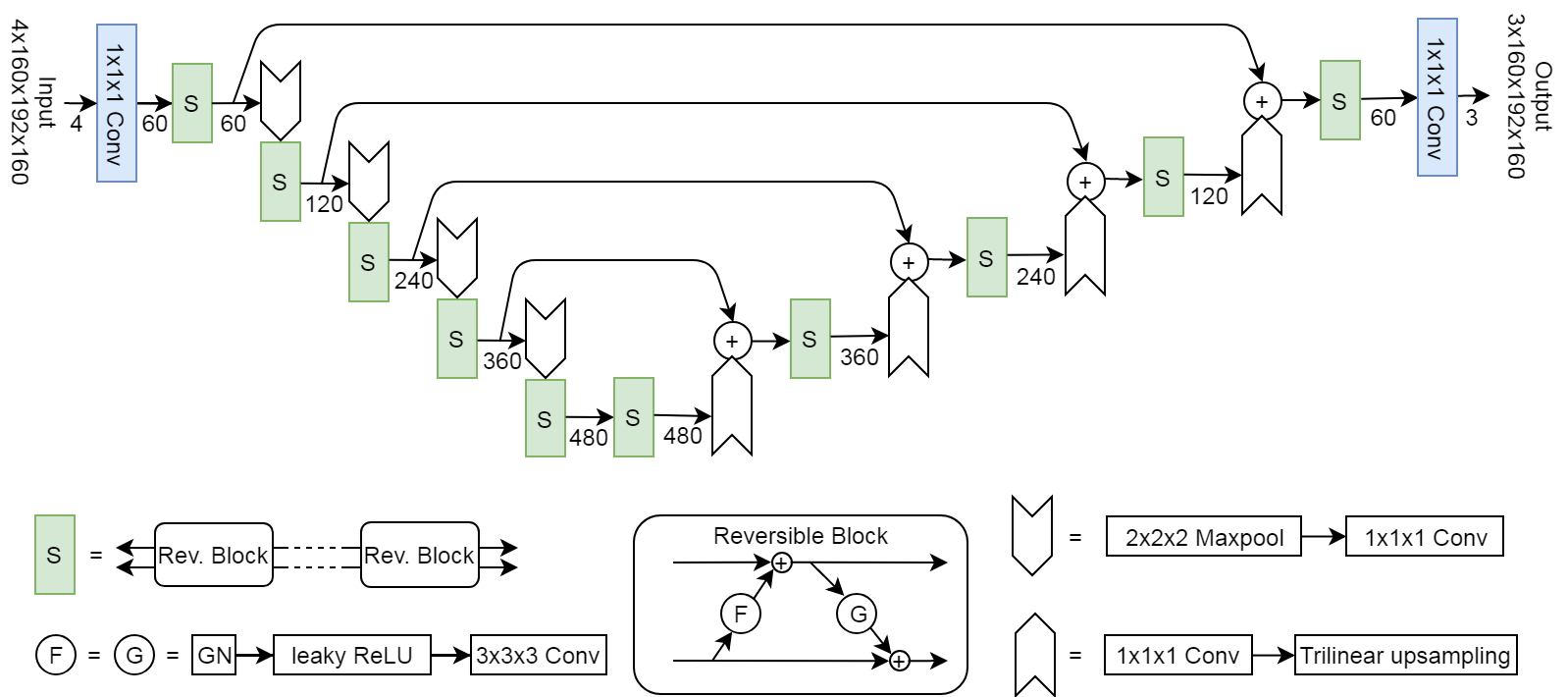}
	\caption{Partially reversible architecture}
	\label{fig:reversible_architecture}
\end{figure}

As mentioned before, a key constraint for a reversible block is that the number of input units must be equal to the number of output units. There are two issues to overcome when adapting reversible blocks in a U-Net architecture. First, spatial downsampling and upsampling which are an integral part of the U-Net \cite{10.1007/978-3-319-24574-4_28} typically change the number of output units. Secondly, the branching and merging nature of the U-Net makes it impractical to model the entire network as a single reversible sequence. While it is possible to create a reversible spatial downsampling operation as demonstrated by Jacobsen et al.\ \cite{jacobsen2018irevnet}, in 3D networks it leads to a prohibitively high number of channels on the lower resolution levels ($d^3$-fold increase for $d$-fold downsampling) and does not solve the problem of the branching and merging network structure. We therefore forgo the idea of a fully reversible architecture and employ a single reversible sequence per resolution level in both the encoder and decoder while using traditional non-reversible operations for down- and upsampling as well as the skip connections. This partially reversible architecture still realizes large memory savings over a traditional U-Net because the activations only need to be saved at the end of each reversible sequence and for the non-reversible components. It also retains the possibility to make the network (almost) arbitrarily deep by increasing the number of reversible blocks in any reversible sequence. This leads to a family of reversible architectures whose members effectively have very similar memory requirements while substantially varying in depth. Figure~\ref{fig:reversible_architecture} shows our architecture.

\subsection{Non-reversible baseline equivalent}
For a fair comparison, we construct a non-reversible baseline with the same number of parameters as the partially reversible architecture with only one reversible block per reversible sequence. To achieve this, we make the following changes:
\begin{itemize}
\item{Replace each reversible block with layers GN--LeakyReLU--Conv--GN--LeakyReLU--Conv, which is effectively a concatenation of $F$ and $G$}
\item{Change the number of channels on the different resolution levels to [30, 60, 120, 240, 480] following the argumentation of Gomez et al. in \ \cite{NIPS2017_6816}. The smaller number of channels on the higher resolution levels are justified by the fact that channels are partitioned into two for the reversible architectures.}
\item{1x1x1 convolutions are not needed to change the number of channels. We therefore omit them.}
\end{itemize}
The non-reversible baseline and the reversible architecture with a single reversible block per reversible sequence both have approximately 12.5M parameters.

\subsection{Memory analysis}
The bulk of the memory consumption when training a non-reversible neural network can be attributed to three categories. Firstly, the activations of the layers ($M_{Al}$). Secondly, the network parameters and memory related to them, such as per-parameter state of some optimizers such as Adam ($M_{Pl}$). Finally, the derivatives of the activations during back-propagation ($M_{Dl}$). 
$M_{Al}$, $M_{Pl}$ and $M_{Dl}$ stand for the respective memory requirement of the $l$-th layer. The tensors in the first two categories need to be allocated for the entire forward- and backward pass. In contrast, memory allocated for the derivatives of the activations can be immediately freed after back-propagation has processed all direct predecessors of a layer. Assuming a non-branching network, the total memory requirement for training can therefore be expressed with the following equation:
\begin{equation}
M_{Total_{NonRev}} = \sum_{l \in Layers} (M_{Al} + M_{Pl}) + \max_{l \in Layers} M_{Dl}
\end{equation}
For a partially reversible architecture, we distinguish between activations for the non-reversible layers ($M_{Nl}$) and the activations at the end of each reversible sequence ($M_{Si}$). We still need the parameter related memory ($M_{Pl}$). The backpropagation of a reversible block is complex and needs additional memory over just the derivatives of the activations ($M_{Bl}$). Again assuming a non-branching network, the total memory needed for a partially reversible architecture can be calculated with the following equation:
\begin{equation}
M_{Total_{PRev}} = \sum_{l \in {NonRevL.}} M_{Nl} + \sum_{i \in RSeq} M_{Si} + \sum_{l \in Layers}  M_{Pl} + \max_{l \in Layers} M_{Bl}
\end{equation}
Each reversible sequence replaces several non-reversible layers, leading to a lower total memory consumption. Furthermore, making partially reversible networks deeper by adding more blocks to any reversible sequence only grows the number of $M_{Pl}$-terms, whereas a non-reversible network also needs additional $M_{Al}$-terms for saving the activations. This comes at the expense of $M_{Bl}$ being larger than $M_{Dl}$ due to the more complex backpropagation process. However, we only need to account for the maximum $M_{Bl}$, therefore the drawback is easily outweighed by having to save fewer activations.

\subsection{Training procedure}
We demonstrate the advantages of the reversible neural network on the BraTS 2018 dataset \cite{6975210}. We implemented the reversible block and the reversible sequence using PyTorch, seamlessly integrating them into the autograd framework. The result is RevTorch\footnote{\url{https://github.com/RobinBruegger/RevTorch}}, an open source library that can be used to create (partially) reversible neural networks. The code used to conduct the experiments in this paper is public as well\footnote{\url{https://github.com/RobinBruegger/PartiallyReversibleUnet}}.

\subsubsection{Preprocessing}
We standardize each image individually to have zero mean and unit variance (based on non-zero voxels only). To avoid initial bias of the network towards one modality, we standardize each modality independently. We set all voxels outside of the brain to zero. We also crop all images to 160x192x160 which fits all but a single brain.

\subsubsection{Augmentation}
We use common training data augmentation strategies including random rotation, random scaling, random elastic deformations, random flips and small shifts in intensity. Because the z-axis of the BraTS dataset has been interpolated, we avoid any augmentation that is not independent of the z-axis. That means we rotate, scale and do elastic deformation only on the planes that are perpendicular to the z-axis. This way, we do not further degrade the image quality.

\subsubsection{Training}
We employ a random data split of 80\%/20\% of the training data provided by the organizers of the BraTS challenge for training and validation. We train using the Adam optimizer with an initial learning rate of $10^{-4}$. The learning rate is decreased by a factor of 5 after 250, 400 and 550 epochs. The batch size is one. We output the nested tumor regions \textit{whole tumor} (WT), \textit{tumor core} (TC) and \textit{enhancing core} (ET) directly. The loss is the unweighted sum of the Dice losses of each of the three regions. Training is stopped once the 30-epoch moving average Dice score on the validation split has not increased for 60 epochs. We regularize using a weight decay of $10^{-5}$. We do not employ any post-processing. For inference, we use the weights which achieved the best Dice score on the validation split. Thanks to the fully convolutional nature of our architecture, we always segment an entire volume in a single pass at test time.

\section{Experiments and Results}
\begin{table}
\caption{Results on the BraTS 2018 validation dataset}
\label{table:results}
\begin{tabular}{|l|c|c|c|c|rrr|rrr|}
\hline
\multicolumn{4}{|c|}{} & Memory & \multicolumn{3}{|c|}{Dice} & \multicolumn{3}{c|}{HD95} \\
Architecture & \#Encoder & \#Decoder & Patchsize & \multicolumn{1}{c|}{(MB)} & \multicolumn{1}{c}{ET} & \multicolumn{1}{c}{WT} & \multicolumn{1}{c|}{TC} & \multicolumn{1}{c}{ET} & \multicolumn{1}{c}{WT} & \multicolumn{1}{c|}{TC}\\
\hline
Baseline & -- & -- & $128^3$ & 6646 & 79.29 & 89.99 & 82.02 & 4.57 & 5.53 & 8.88\\
Reversible & 1 & 1 & $128^3$ & 4138 & 79.02 & 90.24 & 83.92 & 3.19 & 5.33 & 8.33\\
Reversible & 1 & 1 & full & 9436 & 78.99 & 90.39 & 84.40 & 3.16 & 4.67 & 6.39\\
Reversible & 2 & 1 & full & 9529 & 79.66 & 90.68 & 85.47 & 3.35 & 4.64 & 7.38\\
Reversible & 3 & 1 & full & 9620 & 80.33 & 91.01 & 86.21 & 2.58 & 4.58 & 6.84\\
Reversible & 4 & 1 & full & 9713 & 80.56 & 90.61 & 85.71 & 3.35 & 5.61 & 7.83\\
\hline
\multicolumn{3}{|l|}{Isensee et al.\ (2018) baseline} & $128^3$ & -- & 79.59 & 90.80 & 84.32 & 3.12 & 4.79 & 8.16\\
\multicolumn{3}{|l|}{Isensee et al.\ (2018) final submission} & $128^3$ & -- & 80.87 & 91.26 & 86.34 & 2.41 & 4.27 & 6.52\\
\hline
\end{tabular}
\end{table}
We evaluated all models on the online validation dataset provided by the organizers of the BraTS challenge. Note that this dataset is distinct from the validation split described above. Table~\ref{table:results} shows our results. All metrics were computed using the online evaluation platform. \#Encoder and \#Decoder correspond to the number of reversible blocks in each reversible sequence in the encoder and decoder path respectively. The memory consumption was measured using PyTorch's \texttt{max\_memory\_allocated()}. For all performance metrics, the mean of all images in the dataset is reported.

The baseline in row 1 is not reversible and was trained with the patch-based approached mentioned in the introduction to address the memory issue. The reversible equivalent in the 2nd row which was also trained with patches of the same size for direct comparison reduced the memory consumption by more than one third without lowering the performance.

The memory reduction realized by our architecture allows the processing of the whole FOV instead of patches, for which the results are shown in the 3rd row. We believe this is beneficial because it allows to use more context compared to the patch-based approach. While not having improved the Dice scores substantially, it had a positive effect on the Hausdorff metric. As discussed, a key strength of the partially reversible architecture is that we can increase the depth of the network with a very small additional memory requirement coming from the increased number of parameters. Comparing rows 3 to 6 in Table \ref{table:results}, we observe that the memory required when making the network deeper is minimal. In contrast, a non-reversible equivalent of the architecture with four reversible blocks per reversible sequence in the encoder would need over 20GB of memory for the activations alone. We also observe that the segmentation performance increased when going from a single to three reversible blocks per reversible sequence in the encoder before receding again when using four reversible blocks. 

For reference, Table~\ref{table:results} also shows the results reported by Isensee et al.\ \cite{DBLP:journals/corr/abs-1809-10483}. The score most directly comparable to our method is their baseline. It uses the Cross-entropy loss instead of the Dice loss during training, employs test-time-augmentation and is an ensemble of five models trained on different splits of the training data. Their final submission additionally employed post-processing and co-training with additional public and institutional data. Our architecture with three reversible blocks per sequence achieved similar segmentation performance with a single model trained on 80\% of the training data and without any of the additional strategies mentioned above. 

The reported memory savings do come at a cost. We observed a 50\% increase in training time for a reversible architecture over its non-reversible equivalent. However, considering the gains in performance and the opportunities for large-scale architectures, we believe the longer training time is acceptable.

\section{Discussion}
We have demonstrated an extremely memory efficient, partially reversible U-Net architecture for segmentation of volumetric images. We achieve competitive performance compared to current state-of-the art achitectures on the same memory budget. We do this without ensembling, test-time-augmentation, post-processing or co-training with additional data. Applying these techniques to our architecture may further improve performance.
We have demonstrated that our architecture allows arbitrarily deep networks with minimal additional memory requirements. We believe our contribution will allow more researchers to design and investigate large-scale 3D network architectures, even if they do not have access to expensive, highly specialized hardware with massive amounts of memory.
%
%
%
\bibliographystyle{splncs04}
\bibliography{references}
\end{document}